\documentclass[11pt]{article}

\usepackage[preprint]{acl}

\usepackage{times}
\usepackage{latexsym}
\usepackage{multirow}
\usepackage{listings}
\usepackage[T1]{fontenc}

\usepackage[utf8]{inputenc}

\usepackage{microtype}

\usepackage{inconsolata}

\usepackage{graphicx}
\usepackage{booktabs,tabularx,longtable,dcolumn}

\title{Graph-Augmented LLMs for Swiss MP Ideology Prediction}

\author{
Yifei Yuan\textsuperscript{1},
Luis Salamanca\textsuperscript{1},
Sophia Schlosser\textsuperscript{2},
 Laurence Brandenberger\textsuperscript{2}\thanks{Corresponding author.} \\
\textsuperscript{1}Swiss Data Science Center, ETH Zürich, Zürich, Switzerland \\
\textsuperscript{2}Department of Political Science, University of Zürich, Zürich, Switzerland \\
\texttt{yifei.yuan@sdsc.ethz.ch, laurence.brandenberger@ipz.uzh.ch}
}

\begin{document}
\maketitle
\begin{abstract}
Approximating the ideological position of Members of Parliament (MPs) is a fundamental task in political science, helping researchers understand legislative behavior, party alignment, and policy preferences. 
While Large Language Models (LLMs) have shown promising results in estimating MPs’ ideological stances, there are more actors and elements in the parliamentary system, and relations between them, that could provide a wider and more informative picture. 
However, due to the complexity of integrating them in the prediction task, these additional elements are generally ignored. In this work, we propose an LLM framework, \textit{PG-RAG}, that implements a retrieval-augmented generation pipeline: it first queries a political knowledge graph (KG) and then integrates the resulting graph-structured information into the context. This allows for capturing both textual semantics and inter-MP relationships, another relevant information source in any parliamentary system. 
We evaluate the approach on the task of ideology prediction, using data from a Swiss parliamentary dataset. 
When comparing graph-augmented models against several state-of-the-art baselines, the results demonstrate that incorporating this enriched information, which encodes information about different entities and relations, improves prediction performance.
These results help to highlight the value of domain-specific relational information in modeling political behavior.
\end{abstract}

\section{Introduction}


A central question in political science is how to infer the ideological positions of Members of Parliament (MPs) from their observable political behavior~\cite{poole1985spatial, clinton2004statistical}.
Scholars have increasingly used text-based methods to estimate ideology scores from speeches, parliamentary debates, and manifestos, enabling more fine-grained and scalable assessments of MPs' political positions~\cite{laver2003extracting, slapin2008scaling, proksch2010position, lauderdale2016measuring}.

As Large Language Models (LLMs) have demonstrated strong performance across a range of NLP tasks, recent studies have explored their potential for predicting MPs’ ideological stances from textual sources, leveraging their ability to capture semantic nuances and latent political signals embedded in parliamentary speeches~\cite{Liu2022POLITICSPW,Bernardelle2024MappingAI}. Although these approaches have shown reasonable performance, several limitations persist. First of all, they typically treat MPs as independent text generators and ignore the relational structure in parliamentary systems, such as co-sponsorship networks, committee memberships, or party blocs, despite recent work suggesting that such relational information can substantively enrich MPs' representations ~\cite{russo2023helping}. Moreover, LLM-based methods often struggle in low-data settings or when long-term dependencies across multiple parliamentary sessions must be considered, highlighting the need for methods that can jointly leverage textual content and structured relational knowledge~\cite{Huang2024PromptingLL}.

To address these limitations, we propose the integration of graph-structured information to capture ideological alignment and relational influence. By using graph-augmented LLMs, which combine the language understanding capabilities of LLMs with structured relational knowledge captured in graphs, we investigate whether this additional information can improve prediction accuracy. 
Specifically, we propose a \textbf{Political Graph Retrieval-Augmented Generation} (\textit{PG-RAG}) framework, where a graph encoding parliamentary relationships -- such as co-sponsorship links, committee memberships, party affiliations, and ideological clusters -- is queried to retrieve relevant subgraphs for each MP. 
To encode the complexity of the retrieved graphs, comprised by nodes and relations connected to certain MPs and related to specific parliamentary aspects, we explored two approaches. 
First, we leverage the great summarization capabilities of LLMs and prompt the model to first summarize the elements retrieved. 
Second, we explore the ability of the LLM to understand the retrieved graph by providing it as a raw set of nodes and relations.
For each case, the additional context, either the obtained summary or the raw graph, is provided as additional information to a pre-defined prompt, also including some general metadata of the MP. 

The goal of our work is to investigate whether graph-structured information indeed improves ideology prediction and which types of relational signals are most informative. 
To assess this, we compare the described approaches against strong baselines, covering several state-of-the-art LLMs.
To provide a more systematic evaluation, we benchmark zero-shot and few-shot LLM setups, as well as models of different size. 
Our experimental results show that incorporating graph information improves prediction performance, with the effect being particularly pronounced for smaller-scale LLMs. We also observe that LLM-based models struggle when positioning Social Democrats, highlighting directions for future analysis.


\begin{figure*}
\begin{center}
	\includegraphics[width=.95\linewidth]{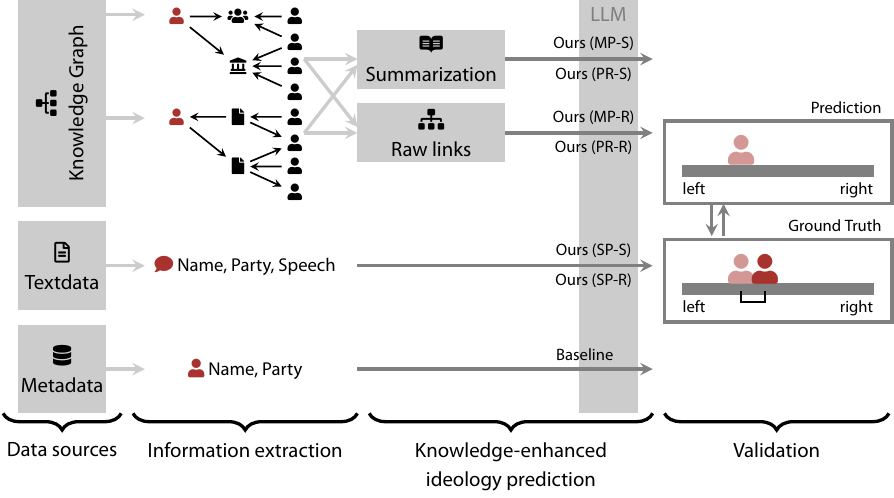}
\end{center}
\caption{The overall framework of our proposed method \textit{PG-RAG}.}\label{fig:framework}
\end{figure*}
\section{Related Work}

\subsection{Ideology Prediction}\label{sec:ideology}
Ideology prediction aims to infer the political or ideological orientation of individuals, groups, or textual content and has been widely studied in the Political Science and NLP domains. 
Early work focused on predicting the ideological leanings of political actors, such as legislators or parties, using legislative speeches, manifestos, or voting records~\cite[e.g.,][]{cox2002measuring, bakker2015measuring, kraft2016embedding, vafa2020text, patil2019roll}. 
Approaches from the Political Science domain relied traditionally on scaling procedures \cite[][]{poole1985spatial, slapin2008scaling, burnham2024semantic}.
In contrast, some NLP-based approaches relied on traditional machine learning models and linguistic features, including SVMs~\cite{SapiroGheiler2019ExaminingPT} and RNNs~\cite{Sinno2022PoliticalIA}, to distinguish ideological positions. 
More recent studies have leveraged pre-trained language models to capture richer contextual representations for ideology prediction across different domains and languages; for instance, ~\citet{Liu2021MitigatingPB} pre-train a Transformer-based language generator to minimize ideological bias in generated text. With the emergence of LLMs, researchers have begun examining whether ideological orientations can be inferred directly from generated or summarized content, as well as how biases present in training data may affect model predictions~\cite{Liu2022POLITICSPW,Bernardelle2024MappingAI,Kim2025LinearRO}.

\subsection{LLMs for Political Tasks}
Recent studies have explored the capabilities of LLMs in political analysis. Prior work shows that LLMs can perform tasks such as political stance detection~\cite{Li2021PStanceAL,Wagner2024ThePO,Pangtey2025LargeLM}, ideology classification~\cite{Haroon2025WhoseSA},  policy analysis~\cite{Chen2025UsingLF}, often achieving performance comparable to or surpassing traditional NLP models. Researchers have also examined the extent to which LLMs encode political biases or ideological patterns in their training data, investigating whether model outputs reflect systematic political preferences or framing effects~\cite{Zhang2025ProbingPI,Kim2025LinearRO,Rettenberger2024AssessingPB}. In addition, several studies evaluate LLMs in political reasoning and multimodal settings, including tasks such as policy debate generation~\cite{dzeparoska2023llm,chuang2025debate}, argument analysis~\cite{Li2025LargeLM}, and political question answering~\cite{Santurkar2023WhoseOD}.
\section{Our Framework}
\subsection{Preliminary}
Given a dataset $D$ consisting of $k$ MP records, $D=\{(i, p_i, g_i, l_i)\}_{i=1}^{k}$, 
where $p_i$ denotes the party of the $i$-th MP, $g_i$ denotes the corresponding party group, 
and $l_i$ represents the ideology score, the ideology prediction task aims to learn a 
function $F$ that maps an specific MP $i$,  and its party and group information $(i, p_i, g_i)$ to the corresponding 
ideology score $l_i$.
\begin{equation}
    l_i = F(i, p_i,g_i)
\end{equation}
The information of the MP encoded in $i$ can simply contain personal and demographic data, such as age and education, which can already support the task of ideology detection. 
However, more complex information can be additionally provided. 
In the following sections, we discuss the proposed methodology \textit{PG-RAG} to further enrich $i$ with information queried from a Political KG.

\subsection{PG-RAG Method}

As shown in Figure~\ref{fig:framework}, we propose a RAG-inspired methodology that leverages information extracted from a political knowledge graph (KG) to address the task of ideology prediction. Specifically, the approach uses the information contained in a (1) \textbf{political KG}, on which it performs (2) \textbf{subgraph extraction}, to finally carry out (3) \textbf{knowledge-enhanced ideology prediction}. The following sections detail these steps. 

\subsubsection{Political KG}
The Political KG utilized is built using the information extracted from the Bulletin of the Swiss Parliament, as detailed in \cite[]{salamanca2024processing}. The schema implemented by this KG aims at encoding the policy-making process, from the moment a pursuit text is proposed by a committee, to all the discussions occurring in the parliament chambers related to it. This is captured through entities such as \texttt{Pursuit} and \texttt{Speech}, with relations encoding temporal dependencies. Furthermore, rich metadata, related to the MPs, Parties, etc., is additionally integrated into the graph, providing further context. A subset of this Political KG, which corresponds to the legislative periods 48th to 51st, ranging from 2007 to 2023, is available at \cite[]{brandenberger2024zenodo}, with further details on the KG structure and its usage. 

\subsubsection{Subgraph extraction}
Due to the large size and complexity of the KG, we decided to define meaningful subgraphs that can be queried independently when generating additional context for the prediction tasks. Each of these subgraphs comprised a subset of entities and the relations connecting them, linked to a specific parliamentary process. Now, given an MP record, we first match it to the corresponding \texttt{Person} node in the KG. Starting from this MP node, the three subgraphs explored are defined according to the following paradigms:
\begin{itemize}
    \item \textbf{Speech-centric (SP)}: We assume that an MP's speeches and legislative activities provide important signals of their ideological position. Therefore, we collect all speeches linked to the MP \texttt{Person} node, through the relation \texttt{gives}. This serves as an important \textbf{baseline} for the other two scenarios, as the retrieved subgraph contains only a single relation and is purely textual.

    \item \textbf{MP-centric (MP)}: In this setting, we extract a subgraph that captures the structural and institutional relationships surrounding the MP (see Figure \ref{fig:mp}). The extracted subgraph includes entities representing the MP’s political affiliations and institutional roles, such as the \texttt{Party}, \texttt{Parliamentary Group}, \texttt{Committee}, and \texttt{Chamber} to which the MP belongs or is elected. In addition, the subgraph incorporates contextual entities describing administrative and geographic connections, such as the represented \texttt{Canton}.    
    \item \textbf{Pursuit-centric (PR)}: The pursuit-centric paradigm focuses on the legislative activities initiated or supported by the MP. Specifically, starting from the target MP node, we retrieve all \texttt{Pursuit} entities that are sponsored or co-sponsored by the MP (see Figure \ref{fig:pr}). These pursuits represent legislative proposals or initiatives that reflect the MP's policy interests and political priorities.
\end{itemize}

Above, the subgraphs are ordered by their complexity. First, in the SP subgraph, the only relevant information is the textual data contained in the \texttt{Speech} node. Hence, this approach is similar to recent methods relying on the semantics of textual data. On the contrary, the MP subgraph captures a true graph structure by querying different entity types within the 1-hop vicinity of the \texttt{Person} node. Finally, the PR subgraph increases the complexity by enabling 2-hop extraction, as well as entities connected by different relation types. The specific queries used to parse the graph are provided in the Appendix \ref{sec:query}.

\subsubsection{Knowledge-enhanced ideology prediction}
Given the extracted subgraph, we need to generate a suitable representation that can be provided as additional context to the LLMs, aiming at improving the ideology prediction task. We propose the following two approaches to encode the subgraph:
\begin{itemize}
    \item \textbf{Summarization (S)}: Motivated by \cite{Zhao2023GraphTextGR}, we first serialize the subgraph into natural language sentences. Specifically, each triplet in the knowledge graph is converted into a textual statement that describes the relationship between entities. These serialized statements collectively form a structured textual representation of the MP's political context, including institutional affiliations, legislative activities, and other relevant relations captured in the subgraph. We use GPT-5 to summarize the subgraph, using the prompts presented in Appendix \ref{sec:sum_prompt}.
    \item \textbf{Raw-Graph (R)}: The graph is provided as retrieved from Neo4J \cite{neo4j}. The JSON formatting encodes the connected nodes as sub-elements, indicating explicitly the relationship type and the different attributes' values. An example is provided in Appendix \ref{sec:subgraph}.
\end{itemize}

Specifically, SP-S provides a summarization of all the queried speeches. In contrast, SP-R presents a subset of them independently. Similarly, (MP/PR)-(S/R) encode truly relational information between different entity types, either serialized into text, or provided as a raw-graph. We report the performance of these 6 different methods accordingly in the following section.

\begin{figure}[t]
\begin{center}
	\includegraphics[width=1\linewidth]{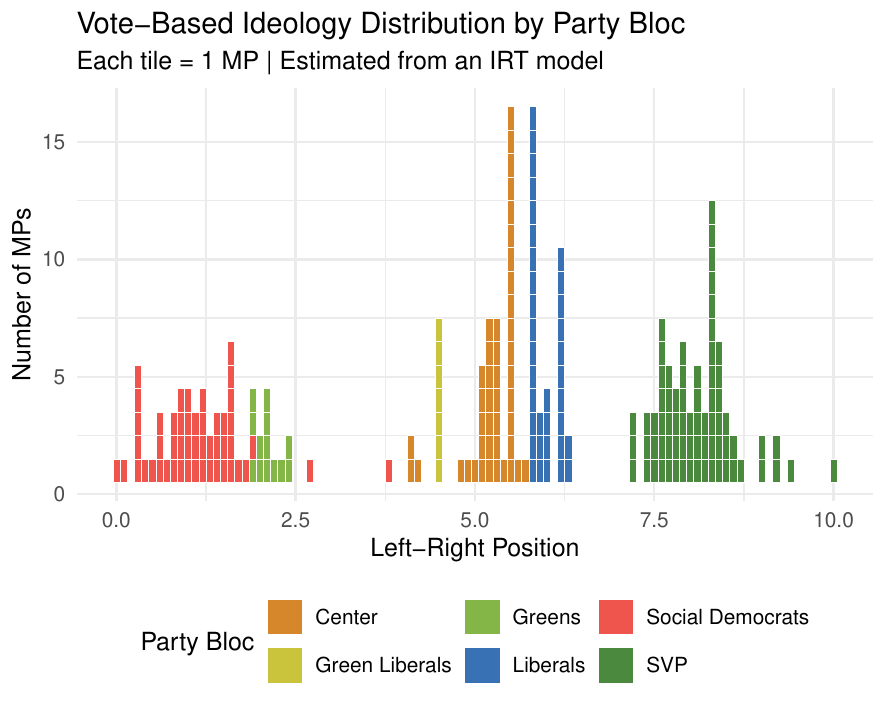}
\end{center}
\caption{Vote-based ideology scores of Swiss members of the National Council}\label{fig_votebasedIdeology}
\end{figure}
\section{Experiments}
\subsection{Experimental Setup}

\textbf{Dataset. }
We collect a dataset from the Swiss National Council, the lower chamber of the Swiss Federal Assembly, comprising 225 unique members of parliament (MPs) during the 50th legislative period (2015-2019). The number exceeds the 200 seats ($N=200$) because some MPs left and were replaced over the four-year period.
We compare ideology predictions for these MPs to vote-based scaled estimates.
These vote-based estimates stem from a random sample of 1000 votes recorded in the Swiss National Council. 
All votes in the National Council are recorded electronically and include points of order votes, standard votes on proposals as well as final votes.
The voting data is provided by the Swiss Parliament\footnote{\url{https://www.parlament.ch/de/ratsbetrieb/abstimmungen/abstimmungs-datenbank-nr}} and incorporated into the DemocraSci KG \cite[][]{brandenberger2026democrasci}.
We use a widely-used dimensional-reduction technique, a two-parameter Item Response Theory model, to estimate a one-dimensional model, as per the standard approach applied in Political Science (for a methodological discussion of scaling procedures, see \citealt[][]{cox2002measuring, cai2016item, bailey2018two}, based on early scaling techniques developed by \citealt[][]{poole1985spatial}). 
We use the \texttt{mirt}-package \cite[][]{chalmers2012mirt} in the Statistical Environment \texttt{R} to estimate vote-based ideology scores.
Figure \ref{fig_votebasedIdeology} shows the stacked distribution of ideology scores (one-dimensional, commonly interpreted as left-right ideological positions).

\begin{table}[t]
    \centering
    \small
    \begin{tabular}{lcccc}
    \toprule
    Method & MAE $\downarrow$&  MSE $\downarrow$ & RMSE $\downarrow$ & RC $\uparrow$ \\
    \midrule
      GM & 2.29  & 7.44 & 2.73 & -   \\
      PM  & 0.30 & 0.20 & 0.44 & 0.97 \\
      PBM & 0.33 & 0.23 & 0.48 & 0.97  \\
    \midrule
    \textbf{zero-shot} \\
    GPT-5 & 0.75 & 1.06 & 1.03 & 0.94 \\
    Qwen3-8B & 1.20 & 2.32& 1.52 & 0.86\\
    Qwen3-32B & 1.11 & 2.03 & 1.43 & 0.86\\
    Apertus-8B & 2.01 & 4.58 & 2.14 & 0.77 \\
    PG-RAG (MP-S) & \textbf{0.72} & 0.77 & 0.88 & 0.94 \\
    PG-RAG (MP-R) & 0.73 & \textbf{0.74} & \textbf{0.86} & \textbf{0.94} \\
    \midrule
    \textbf{few-shot} \\
     GPT-5 & 0.61 & 0.61 & 0.78 & 0.94\\
    Qwen3-8B & 0.88 & 1.57& 1.25 & 0.90\\
    Qwen3-32B & 0.87 & 1.45 & 1.21 & 0.90\\
    Apertus-8B & 2.47 & 7.91 & 2.81 & 0.43\\
    PG-RAG (MP-S) & \textbf{0.58} & \textbf{0.58} & \textbf{0.76} & \textbf{0.94}\\
    PG-RAG (MP-R) & 0.61 & 0.60 & 0.78 & 0.94 \\
        \bottomrule

    \end{tabular}
    \caption{Main results comparing all baseline methods to our best performing models, those using the MP-centric subgraph. RC represents Ranking Correlation.}
    \label{tab:main}
\end{table}

\begin{table}[]
    \centering
    \small
    \begin{tabular}{cccccc}
    \toprule
    Subgraph & Enc.  & MAE &  MSE  & RMSE & RC  \\
    \midrule
    \textbf{zero-shot} \\
    MP & S & \textbf{0.72} & 0.77 & 0.88 & 0.94 \\
    SP & S & 0.73 & 0.78 & 0.88 & 0.93 \\
    PR & S & 0.83 & 1.03 & 1.02 & 0.93 \\
    MP &R & 0.73 & \textbf{0.74} & \textbf{0.86} & \textbf{0.94} \\
    SP& R& 0.76 & 0.78 & 0.89 & 0.94 \\
    PR & R & 0.79 & 0.89 & 0.94 & 0.93\\
    \midrule
    \textbf{few-shot} \\
    MP & S & \textbf{0.58} & \textbf{0.58} & \textbf{0.76} & \textbf{0.94}\\
    SP & S & 0.60 & 0.70 & 0.84 & 0.93\\
     PR & S & 0.68 & 0.83 & 0.91 & 0.94 \\
     MP & R & 0.61 & 0.60 & 0.78 & 0.94 \\
    SP & R & 0.62 & 0.59 & 0.76 & 0.94 \\
     PR & R & 0.62 & 0.60 & 0.77 & 0.94\\
    \bottomrule
    \end{tabular}
    \caption{Exhaustive results for all variants of PG-RAG, using different subgraphs and context encoding methods. It is important to highlight that the SP method does not provide a graph per-se, but rather a summarization of all speeches (SP-S), or a subset of complete speeches (SP-R).}
    \label{tab:extensive}
\end{table}

\noindent\textbf{Evaluation Metrics.} 
We evaluate the ideology prediction performance using the following evaluation metrics: (1) \textbf{Regression Metrics}: Since the task involves predicting continuous ideology scores, we measure prediction accuracy using Mean Absolute Error (MAE), Mean Squared Error (MSE), and Regular Mean Squared Error (RMSE). (2) \textbf{Ranking Metrics}: We also assess whether the model preserves the relative ordering of ideological positions. For this purpose, we employ Spearman's rank correlation ($\rho$).
These sets of metrics provide complementary perspectives, helping to reach more insightful and interpretable results.

\noindent\textbf{Compared Methods.} We compare our method against several baselines, including: (1) \textbf{Naive baselines}: Global Mean (GM), which assigns each MP the overall mean ideology score; Party Mean (PM), which assigns each MP the average ideology score of their party; and Party Bloc Mean (PBM), which assigns each MP the mean ideology score of their party bloc and can be considered an \textbf{upper bound} for this category of methods.
(2) LLM-based methods: We also compare our method with several state-of-the-art LLMs, including GPT-5~\cite{Singh2025OpenAIGS}, the Swiss LLM Apertus~\cite{apertus2025apertus} (8B version), Qwen3-8B and Qwen3-32B-AWQ~\cite{Yang2025Qwen3TR}, under both zero-shot and few-shot settings. For the few-shot setup, we randomly select three examples from the vote-based dataset to serve as in-context demonstrations (prompts see Appendix \ref{sec:zs_prompt} and \ref{sec:fs_prompt}). For all of them, we use the default parameter settings. 

\subsection{Experimental Results}
Table \ref{tab:main} reports the performance of different methods on the ideology prediction task. Several observations can be made: among LLM-based baselines, GPT-5 achieves the best zero-shot performance (MAE = 0.75, Rank Corr = 0.94), while Qwen3-32B performs moderately well. In contrast, Apertus-8B shows substantially weaker performance, suggesting that general-purpose LLMs struggle to infer ideology reliably without structured signals.
When incorporating graph-derived summary knowledge, our approach improves prediction accuracy. In the zero-shot setting, the MP-centric subgraph  reduces MAE to 0.72, while maintaining the same ranking correlation. Furthermore, the MSE is significantly reduced from 1.06 to 0.74 for the PG-RAG (MP-R) case, which demonstrates how our approach is capable of reducing the prediction error even in cases where GPT-5 deviates substantially from the ground truth value. For the few-shot scenario, i.e., when we provide in-context MP examples in the prompt, with their associated metadata and ideology score, the results are more on par. Still, the PG-RAG (MP-S) approach provides some slight improvement. Nevertheless, it is important to highlight that, during our experiments, we noticed a really brittle behavior of the few-shot approach, and adding more examples did not always lead to better results. 

Overall, these results demonstrate that injecting structured knowledge distilled from political graphs into LLM prompts substantially enhances ideology prediction, enabling LLMs to better capture ideological ordering among MPs. In particular, the improvements in MSE for the zero-shot scenario allow to demonstrate how the proposed methods can help recover from predictions that deviate substantially from the ground truth value.

In Table \ref{tab:extensive}, we provide results for all the subgraphs queried and used as context. As discussed before, both SP cases resemble  previous approaches in which only some textual input is provided as context.  In all scenarios, the MP-centric subgraph provides the best results. We believe this is because the MP-centric subgraph provides rich complementary information, such as committees and chambers, while maintaining a moderate context size. On the contrary, the results for the Pursuit-centric subgraph present a degraded performance, likely related to its larger size and complexity, which the LLM still falls short in correctly leveraging. We added more detailed study in Appendix~\ref{sec:variants}.
\section{Extensive Analysis}
\subsection{Party-wise Analysis}
\label{analysis:party-wise}
\begin{figure}
\begin{center}
	\includegraphics[width=1\linewidth]{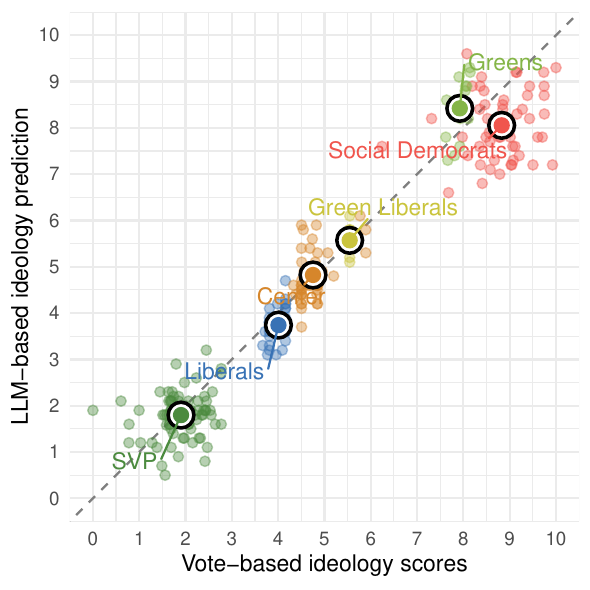}
\end{center}
\caption{Scatterplot of the best LLM prediction scores (MP-S) vs. the vote-based ideology scores. MPs are colored by party blocs and within-bloc averages are highlighted (black-lined circles).}\label{fig_scatterplot}
\end{figure}

Figure \ref{fig_scatterplot} plots the best LLM-based ideology prediction (MP-S) against the vote-based ideology scores. 
MPs are colored by party blocs (with representatives of the Christian and Conservative Democratic Parties merged into the Center bloc). 
Overall, the two ideology scores correlate at a score of $r = 0.963$ (Pearson's, $t = 53.904$, 223 degrees of freedom, $p-value < 0.0001$).
The three smaller centric party blocs all show strong coherence in their ideology predictions.
The two pole party blocs (dominated by SVP and SP members) show more dispersion around the diagonal, indicating that the LLM sometimes predicts the MPs to be more left or right-leaning than would be expected from their voting behavior.
MPs from the Green party bloc (represented by members of the Green party and affiliate communist parties) are generally predicted to be more left by the LLM.
This more left-leaning prediction stems from the fact that the Green party is often ideologically positioned to the left of the Social Democrats and is renowned for collaborating with the Swiss communist parties (who inhabit the left-extreme position) \cite[][]{ladner2019switzerland, ladner2012switzerland, hug2007left, jolly2022chapel}.
However, in the 50th legislative period, the Greens have often voted along left ideological positions and deviated towards the center in order to strengthen their alliances to left-leaning centric members. 
This has brought the MPs from the green party away from the extreme-left position in terms of their voting behavior.
The strongest deviation in the LLM prediction stems from the Social Democrats.
Here, two factors are at work.
First, the LLM judges Social Democrats to be more right-leaning in their ideology than they present in their voting behavior. 
Second, the LLM does not deal with the within-party diversity as well as it does with other party blocs. 
It is worth investigating whether this dispersion stems from the fact that the LLM is unsure where to place these MPs, or whether the signals from their affiliations are different to their signals from their voting behavior. 
The latter could possibly stem from increased party discipline in voting, however, then the ideological placements of Social Democrats from these votes would be more unified.

\begin{figure}
\begin{center}
	\includegraphics[width=1\linewidth]{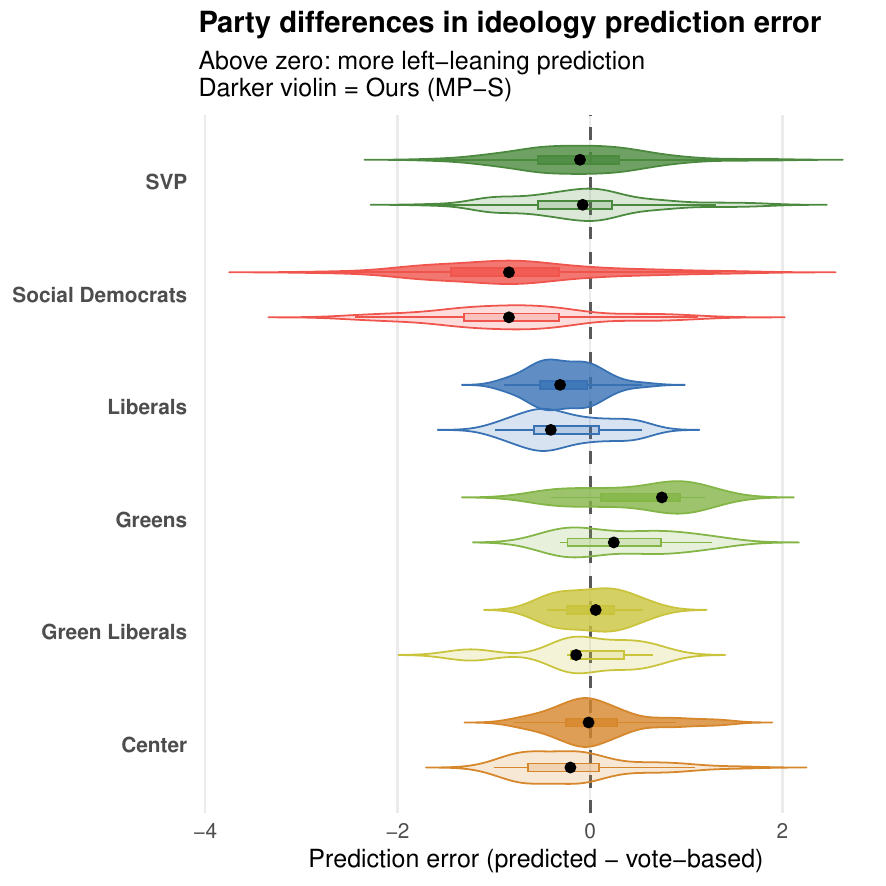}
\end{center}
\caption{Prediction error by party blocs. The dark violins represent PG-RAG (MP-S) model, the lighter violins show the GPT-5 model.}\label{fig_partydifferences}
\end{figure}

Next, we compare the model predictions from GPT-5 (few-shot) and PG-RAG (MP-S) by party bloc in order to see where the additional graph information has helped improve model predictions. 
Figure \ref{fig_partydifferences} presents prediction errors. 
For some party blocs, the additional graph information helps contract the predictions around the true (vote-based) values. 
This is the case for the Liberals, the Green Liberals as well as for the Center party bloc. 
However, for others, the raw GPT-5 predictions are closer to the vote-based ideologies (Greens, SVP). 
This should not necessarily be interpreted as a failure of our model, but rather that the inclusion of additional MP information has shifted the ideology predictions. 
It is well-known that voting behavior is not the only ideological indicator in legislative studies \cite[][]{snyder2000estimating, rheault2020word, barber2022comparing} and, as we have indicated, it is based on legislative behavior that is biased in and of itself \cite[e.g.,][]{carrubba2006off, carrubba2008legislative, hug2010selection}.
As such, it would be interesting to study in larger and more varied datasets whether the LLM-based and graph-enhanced ideology predictions reflect a more nuanced ideology placement that encompasses MPs' legislative behavior outside voting.

Figure \ref{fig_partydifferences} also shows the difficulties the LLM-based models face when placing Social Democrats.
Here, the prediction errors span the largest range in both models, indicating that both the raw GPT-5 as well as the graph-enhanced predictions are generally more right-leaning for Social Democrats than their voting behavior would suggest. 
The reason for this distortion needs to be explored further: is it based on heterogeneous actions by these MPs that make them difficult to pinpoint, or is it an inherent bias in LLMs to bias Social Democrats more towards the right?

\subsection{Backbone Analysis} 
\begin{figure}
    \centering
    \includegraphics[width=1.0\linewidth]{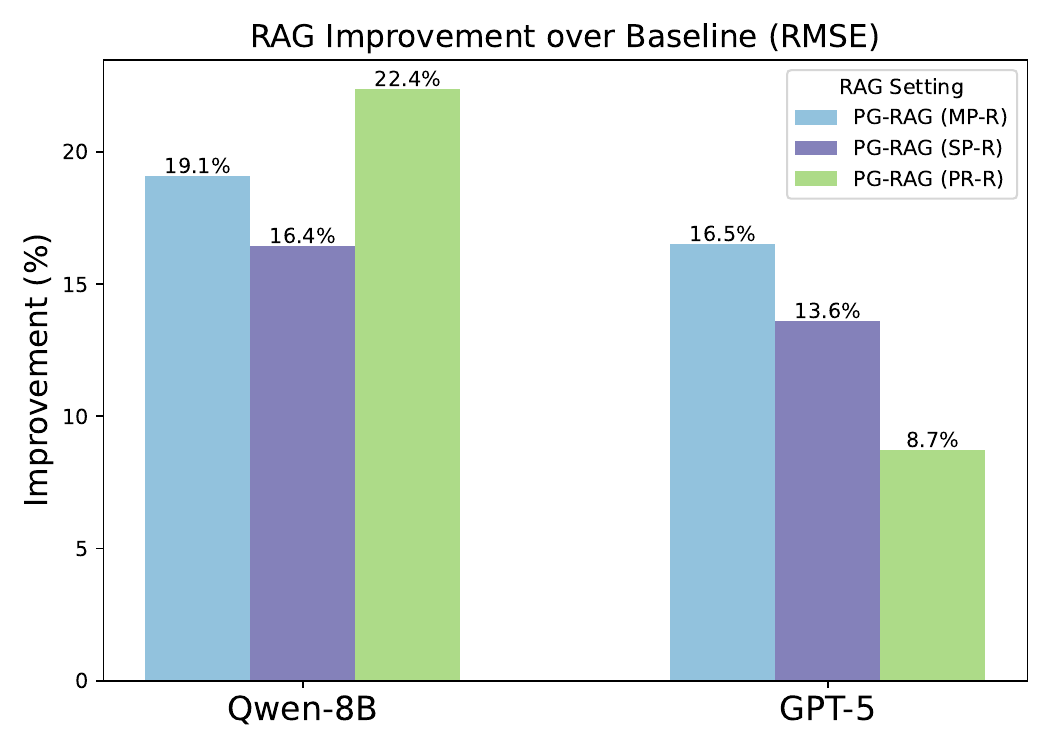}
    \caption{Performance improvements from RAG over the non-RAG baseline for Qwen-8B and GPT-5. }
    \label{fig:sec5.2}
\end{figure}
\begin{table*}[]
    \centering
    \small
    \begin{tabularx}{\textwidth}{llXlcc}
    \toprule
     Person    &  Party (Party Bloc)  & Summary of the MP-centric subgraph & GT & GPT-5 & PG-RAG\\
     \midrule
     MP-A & LDP (Liberals)  &    \setlength{\baselineskip}{7pt} {\scriptsize
 MP-A (b. [YEAR], [CITY]) is a National Council (NR) member for [CANTON]. He belongs to the Liberal Democratic Party (LDP) and sits with the FDP-Liberale parliamentary group (RL), the liberal bloc in Swiss politics. He serves on the [COMMITTEE SEAT]. The LDP/FDP family in Switzerland generally emphasizes a market-oriented economy, individual liberties, and strong support for education, research, and cultural policy.}
 & 4.2 & 4.6 & 4.2 \\ 
MP-B & FDP-Liberale (Liberals) &   \setlength{\baselineskip}{7pt} {\scriptsize MP-B (b. [YEAR], [CITY]) is a National Council (NR) member representing the canton of [CANTON]. He is a member of FDP. Die Liberalen and sits in the FDP-Liberal parliamentary group (formerly Freisinnig-demokratische Fraktion). The FDP in Switzerland is associated with market-oriented economic policy, competitiveness and individual freedoms. He serves on the [COMMITTEE SEAT A] and the [COMMITTEE SEAT B] committees, focusing on these policy areas.}
  & 3.8 & 2.9 & 3.6 \\     
MP-C & SVP (SVP) &   \setlength{\baselineskip}{7pt} {\scriptsize
   MP-C (born [YEAR] in [CITY]; citizen of [CITIZENSHIP]) represents [CANTON]] in the Council of States (SR). He is a member of the Swiss People's Party (SVP) and its parliamentary group (V). His committee work includes the [COMMITTEE SEAT] and an ad hoc committee [COMMITTEE NUMBER]. The SVP is a right-wing, conservative party emphasizing national sovereignty, restrictive immigration policy, lower taxes, and skepticism toward EU integration.}
 & 1.7 & 2.1 & 1.8\\
MP-D & SP (Social Democrats) &     \setlength{\baselineskip}{7pt} {\scriptsize  MP-C (born [YEAR] in [CITY]) is a National Council member for the canton of [CANTON] from the Sozialdemokratische Partei der Schweiz (SP) and sits in the Sozialdemokratische Fraktion. The SP is a center-left social-democratic party advocating social justice, strong public services, labor rights, and progressive social policy. Naef serves on the [COMMITTEE SEAT A] and the [COMMITTEE SEAT B], indicating a focus on international and legal matters. He is a citizen of the city of [CITIZENSHIP].}
 & 9.0 & 6.5 & 7.1\\
\bottomrule
    \end{tabularx}
    \caption{Examples (de-identified) of the MP ideology prediction from GPT-5 and PG-RAG (MP-S). The summaries are generated through GPT-5 using the information queried from the KG, for the MP-centric subgraph. The prompt used to carry out ideology prediction contained the fully identified information.}
    \label{tab:casestudy}
\end{table*}

Figure \ref{fig:sec5.2} shows the RMSE improvement from RAG over the non-RAG baseline for Qwen-8B and GPT-5 under the zero-shot scenario, with the context encoding method as raw subgraph (R). From the figure, we observe that the PG-RAG (MP-R) setting yields the most significant improvement for GPT-5, reducing the error rate by 16.5\%. This suggests that for large-scale models, providing high-density MP-centric graph context is highly effective. Additionally, Qwen-8B consistently shows higher percentage improvements across all RAG settings (ranging from 16.4\% to 22.4\%) compared to GPT-5 (8.7\% to 16.5\%). This indicates that graph-based context provides a more substantial "knowledge boost" to the model with a smaller scale than to the more advanced GPT-5. Interestingly, while the PG-RAG (PR) (Pursuit-based) setting was the strongest for Qwen-8B, it shows the weakest improvements for GPT-5, providing an 8.7\% improvement. This suggests that smaller models depend more on external structured guidance, whereas larger models can internally absorb and reason over the same information. Overall, integrating graph knowledge provides a moderate, stable improvement for both models, acting as a reliable middle-ground strategy.

\subsection{Case Study}
We further demonstrate several examples to understand our model, MP-S case, compared to GPT-5 raw model in Table \ref{tab:casestudy}. We find that across the cases, our model's predictions consistently align more closely with the ground truth than GPT-5. 
For MP-A and MP-B, both members of liberal parties, our prediction is almost identical to the GT.
Similarly, for MP-C, our prediction is nearer to the GT than GPT-5's.
In these cases, our method is better able to leverage the graph and textual context to make accurate predictions. 
In addition, we observe that for politicians from left-leaning parties, such as MP-D, our model predicts a score of 7.1, which is closer to the ground-truth value of 9.0 compared to GPT‑5's prediction of 6.5. This example highlights the challenge of accurately predicting left-leaning MPs, consistent with our findings in Section \ref{analysis:party-wise}. This difficulty arises because left-leaning MPs often have high ideology scores, while LLMs tend to generate more moderate predictions.
Our model partially mitigates this difficulty, producing predictions closer to the ground truth for some of these cases. By leveraging textual summaries from the knowledge graph, which capture rich contextual information, such as committee memberships and policy focus, our model enables better predictions.


\section{Conclusion}
We introduce \textit{PG-RAG}, a RAG-inspired graph-augmented LLM framework for political ideology prediction. Leveraging the data from a Swiss parliamentary knowledge graph,  we explore three subgraph scenarios, speech-centric, MP-centric, and pursuit-centric, with two context encoding methods for each, correspondingly. We then compare our method with several LLMs. Our experiments show that with graph-structured relational data, our approach captures the complex web of inter-MP relationships and parliamentary elements that define legislative behavior. In addition, with graph knowledge, the model shows improved understanding of political tendencies across different parties. These results lay the groundwork for future work on extending to additional parties and enhancing performance on left-leaning parties.

\section*{Limitations}

While our model demonstrates strong overall predictive performance, it still exhibits reduced accuracy when predicting positions of left-leaning parties. This suggests that the model may not fully capture the nuances in the rhetoric or policy preferences characteristic of these groups. Additionally, our current dataset and analysis are limited to a subset of MPs, and extending the model to a broader set of representatives could improve generalizability. However, this expansion is not feasible for Switzerland due to data constraints and the limited availability of annotated parliamentary records.

\section*{Acknowledgments}
This work is supported by the Swiss National Science Foundation (grant 10.003.190, measuring political success).

\bibliography{custom,customLB-donottouch}

\appendix

\section{LLM prompts}
\label{sec:appendix}
\subsection{Zero-shot setting}
\label{sec:zs_prompt}
\begin{lstlisting}
Predict the ideology score of this Swiss MP (0=Right, 10=Left). Important guidelines:
- Base your judgment primarily on the MP's background information.
- Do NOT rely only on the party or bloc label.
- MPs within the same party or bloc can have different ideological positions.
- Choose a precise value (e.g., 3.7 or 6.2).

Name: {MP Name}, Party: {MP Party}, Bloc: {MP Party Block}
Return ONLY the number.
Score:
\end{lstlisting}
\subsection{Few-shot setting}
\label{sec:fs_prompt}
\begin{lstlisting}
You are a political scientist. Below are examples of MPs and their ideology scores (0 = Far-Right, 10 = Far-Left):
Name: {example MP name} | Party: {example MP party} | Party Bloc: {example MP party bloc} | Score: {example MP ideology score}
...

Predict the ideology score of this Swiss MP (0=Right, 10=Left). Important guidelines:
- Base your judgment primarily on the MP's background information.
- Do NOT rely only on the party or bloc label.
- MPs within the same party or bloc can have different ideological positions.
- Choose a precise value (e.g., 3.7 or 6.2).

Name: {MP Name}, Party: {MP Party}, Bloc: {MP Party Bloc}
Return ONLY the number.
Score:
\end{lstlisting}
\subsection{Prompt for subgraph summarization}
\label{sec:sum_prompt}
\begin{lstlisting}
You are a political science expert.

Given structured information about a Member of Parliament (MP), write a concise neutral description of this politician that could help infer their political ideology.

Avoid speculation and keep the description factual. 

MP information:
{mp_context}

Write a short summary (around 500 characters) describing the MP's political background and potential ideological positioning.

\end{lstlisting}
\begin{table*}[t]
\centering
\small
\begin{tabular}{lcccccccccccccc}
\toprule
\multirow{2}{*}{\textbf{Model}} & \multicolumn{4}{c}{\textbf{MP-centric}}& & \multicolumn{4}{c}{\textbf{Speech-centric}} & &\multicolumn{4}{c}{\textbf{Pursuit-centric}} \\
\cline{2-5} \cline{7-10} \cline{12-15}
 & MAE & MSE & RMSE & RC& & MAE & MSE & RMSE & RC & & MAE & MSE & RMSE & RC \\
\midrule
\textbf{zero-shot} \\
PG-RAG (r-10) &  0.75 & 0.81 & 0.90 & 0.94 & & 0.80 & 0.88 & 0.94 & 0.94 & & 0.85 & 0.99 & 1.00 & 0.92\\
PG-RAG (r-50) & 0.77 & 0.83 & 0.91 & 0.94 & &
0.77 & 0.81 & 0.90 & 0.94 & & 0.80 & 0.92 & 0.96 & 0.93\\
PG-RAG (r-100) & 0.79 & 0.85 & 0.92 & 0.93 & & 0.78 & 0.84 & 0.92 & 0.94 & & 0.80 & 0.91 & 0.95 & 0.93\\
\midrule
\textbf{few-shot} \\
PG-RAG (r-10) & 0.55 & 0.55 & 0.74 & 0.94 & &  0.61 & 0.60 & 0.78 & 0.94 & & 0.66 & 0.72 & 0.85 & 0.93\\
PG-RAG (r-50) & 0.60 & 0.61 & 0.78 & 0.94 & & 0.60 & 0.59 & 0.77 &0.94 & & 0.68 & 0.72 & 0.85 & 0.93\\
PG-RAG (r-100) & 0.59 & 0.59 & 0.77 & 0.94 & & 0.61 & 0.61 & 0.78 & 0.93 & & 0.68 & 0.74 & 0.86 & 0.94 
\\
\bottomrule
\end{tabular}
\caption{Performance comparison of ideology prediction under different settings.}
\label{tab:variants}
\end{table*}

\section{Subgraph Demonstration}
\subsection{MP-centric}

\begin{figure}[!hbt]
    \centering
    \includegraphics[width=0.95\linewidth]{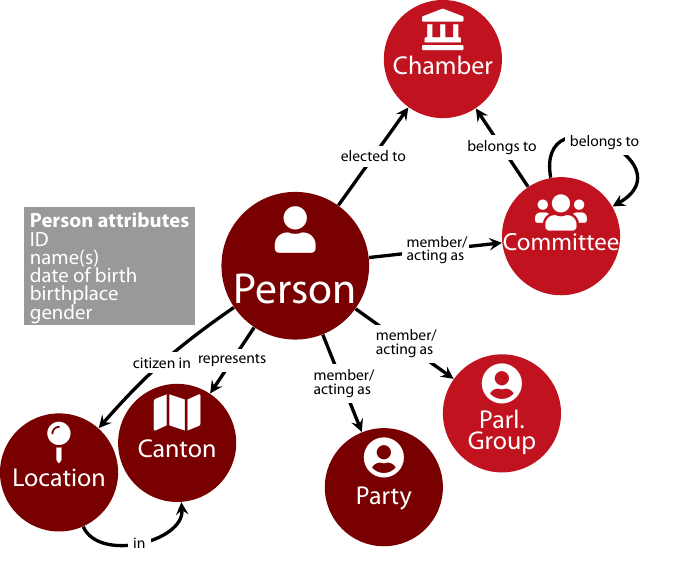}
    \caption{Graph demonstration of the MP-centric subgraph.}
    \label{fig:mp}
\end{figure}

Figure \ref{fig:mp} shows the MP-centric subgraph. 
For the focal MP, we extract the chamber they are elected to, the committees they sit on, the parties and parliamentary groups they belong to, the canton they represent and the city they live in. 
Whenever relations are time-stamped we extract only those relations that are within the 50th legislative period.

\subsection{Pursuit-centric}

\begin{figure}[!hbt]
    \centering
    \includegraphics[width=0.95\linewidth]{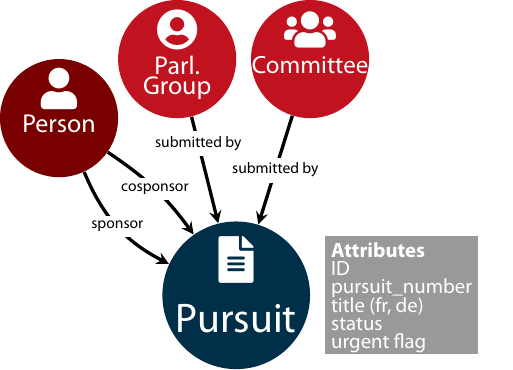}
    \caption{Graph demonstration of the pursuit-centric subgraph.}
    \label{fig:pr}
\end{figure}

Figure \ref{fig:pr} shows the pursuit-centric subgraph. This two-hop subgraph starts with the focal MP and their links to pursuits. These relations either represent sponsorship (i.e., submitting agent) or co-sponsorship (i.e., supporting agent). 
The second hop represents links from the pursuit to other cosponsors as well as to committees or parliamentary groups who can act as sponsors themselves.

\section{Example Subgraph Demonstration}
\label{sec:subgraph}
Here's an example of how our extracted Neo4J subgraph looks in the MP-centric scenario:
\begin{lstlisting}
[
  {
    "csv_uid": "74",
    "original_label": null,
    "person_info": {
      "date_birth": "xxx",
      "uid": "xx",
      "gender": "x",
      "last_name": "xxx",
      "first_name": "xxx"
    },
    "graph_context": [
      {
        "type": "City",
        "rel": "BORN_IN",
        "properties": {
          "post_code": "4000",
          "name": "Basel"
        }
      },
      {
        "type": "City",
        "rel": "CITIZEN_IN",
        "properties": {
          "post_code": "4000",
          "name": "Basel"
        }
      },
    ...
    ]
  }
]
\end{lstlisting}
\section{Graph-parsing Queries}
\label{sec:query}
We provide the queries we used to parse the subgraph under the three scenarios.
\subsection{Speech-centric}
\begin{lstlisting}
MATCH (p:Person {uid: $uid})
OPTIONAL MATCH (p)-[r]-(neighbor:Speech)
WHERE datetime(neighbor.time_end)   >= datetime("2015-11-30T00:00:00")
  AND datetime(neighbor.time_start) <= datetime("2019-12-01T23:59:59")
RETURN 
    properties(p) as p_props,
    labels(neighbor)[0] as n_label, 
    properties(neighbor) as n_props, 
    type(r) as rel_type
\end{lstlisting}
\subsection{MP-centric}
\begin{lstlisting}
 MATCH (p:Person {uid: $uid})-[r]-(neighbor)
WHERE any(label IN labels(neighbor) WHERE label IN [
    'Chamber', 'Committee', 'Party', 'Canton', 
    'Location', 'Parliamentary Group'
])
WITH p, neighbor, r,
     labels(neighbor)[0] AS n_label,
     properties(neighbor) AS n_props,
     type(r) AS rel_type
// Apply constraints for Chamber and Committee
WHERE (n_label = 'Chamber' AND rel_type = 'ELECTED_TO' 
       AND r.date_election >= date("2015-11-30") 
       AND r.date_election <= date("2019-12-01"))
   OR (n_label = 'Committee' 
       AND r.date_joining >= date("2015-11-30") 
       AND r.date_leaving <= date("2019-12-01"))
   OR (n_label <> 'Chamber' AND n_label <> 'Committee')
RETURN DISTINCT properties(p) AS p_props, n_label, n_props, rel_type
\end{lstlisting}
\subsection{Pursuit-centric}
\begin{lstlisting}
MATCH (p:Person {uid: $uid})

OPTIONAL MATCH (p)-[r1]-(n1:Pursuit)

OPTIONAL MATCH (n1)-[rs:SUBMITTED_TO]->()
WHERE rs.date >= date("2015-11-30")
  AND rs.date <= date("2019-12-01")

OPTIONAL MATCH (n1)-[r2:SPONSORS|COSPONSORS]-(n2)
WHERE n2 IS NULL OR  n2 <> p

RETURN DISTINCT
    properties(p) AS p_props,
    labels(n1)[0] AS n1_label,
    properties(n1) AS n1_props,
    type(r1) AS r1_type,
    labels(n2)[0] AS n2_label,
    properties(n2) AS n2_props,
    type(r2) AS r2_type
\end{lstlisting}

\section{Model Variants Comparison}
\label{sec:variants}
To avoid generating a massive context that could instead confound the model, we restrict the raw graph to $N$ nodes. To ensure a correct coverage of the full subgraph, these are randomly selected among all the sub-elements. Table \ref{tab:variants} shows the performance comparison under different setups when we randomly include $N$ nodes with the prompt.
\end{document}